\documentclass[a4paper,fleqn]{cas-dc}

\usepackage[authoryear,longnamesfirst]{natbib}
\usepackage{graphicx}
\usepackage{caption}
\usepackage{subcaption}
\usepackage[ruled,vlined, linesnumbered]{algorithm2e}

\SetKwComment{Comment}{\color{green!50!black}// }{}

\SetKwProg{Function}{function}{}{}

\def\tsc#1{\csdef{#1}{\textsc{\lowercase{#1}}\xspace}}
\tsc{WGM}
\tsc{QE}
\tsc{EP}
\tsc{PMS}
\tsc{BEC}
\tsc{DE}

\begin{document}
\let\WriteBookmarks\relax
\def\floatpagepagefraction{1}
\def\textpagefraction{.001}

\shorttitle{Image Registration of Micro-Ultrasound and Pseudo-Whole Mount Histopathology Images of the Prostate}

\shortauthors{Imran et~al.}

\title [mode = title]{Image Registration of \textit{In Vivo} Micro-Ultrasound and \textit{Ex Vivo} Pseudo-Whole Mount Histopathology Images of the Prostate: A Proof-of-Concept Study}             

\author[1]{Muhammad Imran}
\credit{Data curation, Methodology, Software, Formal analysis, Investigation, Writing - original draft, Visualization}
\author[2]{Brianna Nguyen}
\credit{Data curation, Writing - review \& editing}
\author[3]{Jake Pensa}
\credit{Data curation, Writing - review \& editing} 
\author[4]{Sara M. Falzarano}
\credit{Data curation, Writing - review \& editing}
\author[5]{Anthony E. Sisk}
\credit{Writing - review \& editing}
\author[6]{Muxuan Liang}
\credit{Formal analysis, Writing - review \& editing}
\author[2]{John Michael DiBianco}
\credit{Data curation, Writing - review \& editing}
\author[2]{Li-Ming Su}
\credit{Supervision, Writing - review \& editing}
\author[7]{Yuyin Zhou}
\credit{Supervision, Writing - review \& editing}
\author[8]{Wayne G. Brisbane}
\fnmark[1]
\credit{Conceptualization, Resources, Supervision, Writing - review \& editing, Project administration}
\author[1]{Wei Shao}[
orcid = 0000-0003-4931-4839]
\cormark[1]
\fnmark[1]
\ead{weishao@ufl.edu}
\credit{Conceptualization, Methodology, Software, Resources, Data curation, Writing review \& editing, Supervision, Project administration, Funding acquisition}
\affiliation[1]{organization={Department of Medicine, University of Florida},
    city={Gainesville},
    state={Florida},
    postcode={32608}, 
    country={United States}
    }
\affiliation[2]{organization={Department of Urology, University of Florida},
    city={Gainesville},
    state={Florida},
    postcode={32608}, 
    country={United States}
}
\affiliation[3]{organization={Department of  Bioengineering, University of California},
    city={Los Angeles},
    state={California},
    postcode={90095}, 
    country={United States}
    }
    \affiliation[4]{organization={Department of Pathology, Immunology and Laboratory Medicine, University of Florida},
    city={Gainesville},
    state={Florida},
    postcode={32610}, 
    country={United States}
    }
    \affiliation[5]{organization={Department of  Pathology, University of California},
    city={Los Angeles},
    state={California},
    postcode={90095}, 
    country={United States}
    }
    \affiliation[6]{organization={Department of Biostatistics, University of Florida},
   city={Gainesville},
    state={Florida},
    postcode={32608}, 
    country={United States}
    }

    \affiliation[7]{organization={Department of Computer Science and Engineering, University of California},
    city={Santa Cruz},
    state={California},
    postcode={95064}, 
    country={United States}
    }
    \affiliation[8]{organization={Department of  Urology, University of California},
    city={Los Angeles},
    state={California},
    postcode={90095}, 
    country={United States}
    }

\cortext[cor1]{Corresponding author}
\fntext[fn1]{Equal contribution as senior author}

\begin{abstract}
Early diagnosis of prostate cancer significantly improves a patient's 5-year survival rate.  Biopsy of small prostate tumors is improved with image-guidance. MRI-ultrasound (micro-US) fusion-guided biopsy is sensitive to small tumors but is underutilized due to the high cost of MRI and fusion equipment. 
Micro-US, a novel high-resolution ultrasound technology, provides a cost-effective alternative to MRI while delivering comparable diagnostic accuracy.
However, the interpretation of micro-US is challenging due to the subtlety of the grayscale changes that indicate cancer compared with normal tissue.  
This challenge can be  addressed by training urologists with a large dataset of micro-US images containing  ground-truth cancer outlines.  Such a dataset can be mapped from surgical specimens (histopathology) onto micro-US images through image registration.
In this paper, we present a semi-automated pipeline for registering \textit{in vivo} micro-US images with \textit{ex vivo} whole-mount histopathology images.
Our pipeline begins with the reconstruction of pseudo-whole mount histopathology images and a 3-dimensional (3D) micro-US volume.
Each pseudo-whole mount histopathology image is then registered with the corresponding axial micro-US slice using a two-stage approach that estimates an affine transformation followed by a deformable transformation.
We evaluated our registration pipeline using micro-US and histopathology images from 18 patients who underwent radical prostatectomy. 
The results showed a Dice coefficient of 0.94 and a landmark error of 2.7 mm, indicating the accuracy of our registration pipeline. 
This proof-of-concept study demonstrates the feasibility of accurately aligning micro-US and histopathology images. 
To promote transparency and collaboration in research, we will make our code and dataset publicly available.

\end{abstract}




\maketitle

\section{Introduction}
\label{section1}
Prostate cancer is a significant public health concern worldwide. It is currently regarded as the second most prevalent cancer globally, with 1.4 million new cases and 375,304 related deaths reported in 2020 \citep{Ref1, Ref2}. In the United States, the American Cancer Society estimates  there will be 288,300 new prostate cancer cases and 43,700 deaths in 2023 \citep{Ref3}.  The diagnosis of prostate cancer includes prostate biopsy, a procedure performed over 1 million times annually in the United States \cite{loeb2011complications}.  Concerning lesions in the prostate are often first identified on MRI, and subsequently biopsied under ultrasound with the MRI target fused to a live conventional ultrasound (MRI-transrectal ultrasound [TRUS]-guided biopsy).  
  However, the implementation of MRI-TRUS-guided biopsy presents challenges.  First, MRI is expensive and rarely available. Second, the interpretation of MRI requires sub-specialized radiologists.  Finally, the fusion software overlying MRI and ultrasound may produce errors \citep{Ref7}.

Micro-ultrasound (micro-US) is an emerging technology that offers numerous advantages over conventional ultrasound and MRI. 
Operating at a high frequency of 29 MHz, micro-US provides image resolution three to four times higher than conventional ultrasound \citep{Ref9}. 
This high resolution enables real-time visualization of prostate cancer, delivering diagnostic accuracy comparable to MRI \citep{Ref6, Ref7, Ref8, Ref9, Ref10}. 
In addition, micro-US costs approximately one-tenth that of an MRI scan.
However, the interpretation of micro-US is complex due to a lack of familiarity among urologists with the high resolution and subtle acoustic changes that differentiate cancer from normal tissue.  

Overlaying histopathology onto micro-US images will greatly improve prostate cancer care in three ways:  First, by providing annotation of cancer versus benign tissue on micro-US, our work enables the development of diagnostic machine-learning algorithms. Such algorithms can improve the accuracy of micro-US and increase its utilization amongst urologists. Second, our data will enable analysis of which micro-US acoustic features indicate an aggressive cancer phenotype.  Finally, our work will improve tumor volume predictions for visible lesions on micro-US. 

We aimed to perform image registration between micro-US and pseudo-whole mount histopathology images obtained from patients undergoing surgical prostate removal.  The registration of micro-US and histopathology images presents several challenges. First, micro-US images are acquired in the pseudo-sagittal (oblique) plane, while histopathology is processed in the axial plane. 
Second, Histopathology images can undergo substantial shape deformations due to uneven tissue fixation, excision, staining, and staging \citep{Ref17, Ref18}.
While various methods have been proposed to address the challenges of registering prostate MRI and histopathology images \citep{Ref11, Ref17, Ref19, Ref20, Ref21, Ref22,shao2021prosregnet,sood20213d}, none of these techniques can be directly applied to the registration of micro-US and histopathology images.

In this paper, we present a semi-automated pipeline for accurate registration of \textit{in-vivo} micro-US and \textit{ex-vivo} pseudo-whole mount images. Our registration pipeline involves multiple steps to ensure precise alignment between histopathology and micro-US images.
First, we reconstructed pseudo-whole mount histopathology images by stitching histopathology fragments.
Next, we transformed 2-dimensional (2D) micro-US images from oblique planes to the axial plane, creating a 3-dimensional (3D) micro-US volume to match the orientation of histopathology images.
Then, we manually estimated slice correspondences between axial micro-US and pseudo-whole mount histopathology images by visually identifying corresponding prostate features.
Finally, we performed pairwise 2D affine and deformable image registrations between each histopathology and the corresponding axial micro-US slice. The registration process utilized prostate segmentations and the mutual information loss function.
 This paper makes the following major contributions:

\begin{itemize}

\item We developed the first semi-automated approach for registering \textit{in-vivo} micro-US and \textit{ex-vivo} pseudo-whole mount histopathology images.
\item Our image registration approach has achieved a mean landmark error of 2.7 mm, which is comparable to prior MRI-histopathology registration methods. This approach demonstrates the feasibility of accurately registering micro-US and histopathology images acquired at different orientations.
\item To validate our registration approach as a proof-of-concept study, we conducted experiments on a medium-size dataset comprising 18 patients who underwent radical prostatectomy. 
\item By accurately mapping pathologist-annotated prostate cancer outlines from histopathology onto micro-US images, we have created a valuable and unique dataset for developing machine learning methods for prostate cancer detection on micro-US images.
\item We will make our accurately aligned micro-US and histopathology images publicly available, fostering collaboration and advancements in the field of micro-ultrasound and histopathology image registration.
\end{itemize}

\section{Methods}
\label{sec:methodology}
\begin{figure*}[!h]
\centering
\includegraphics[width=0.98\textwidth]{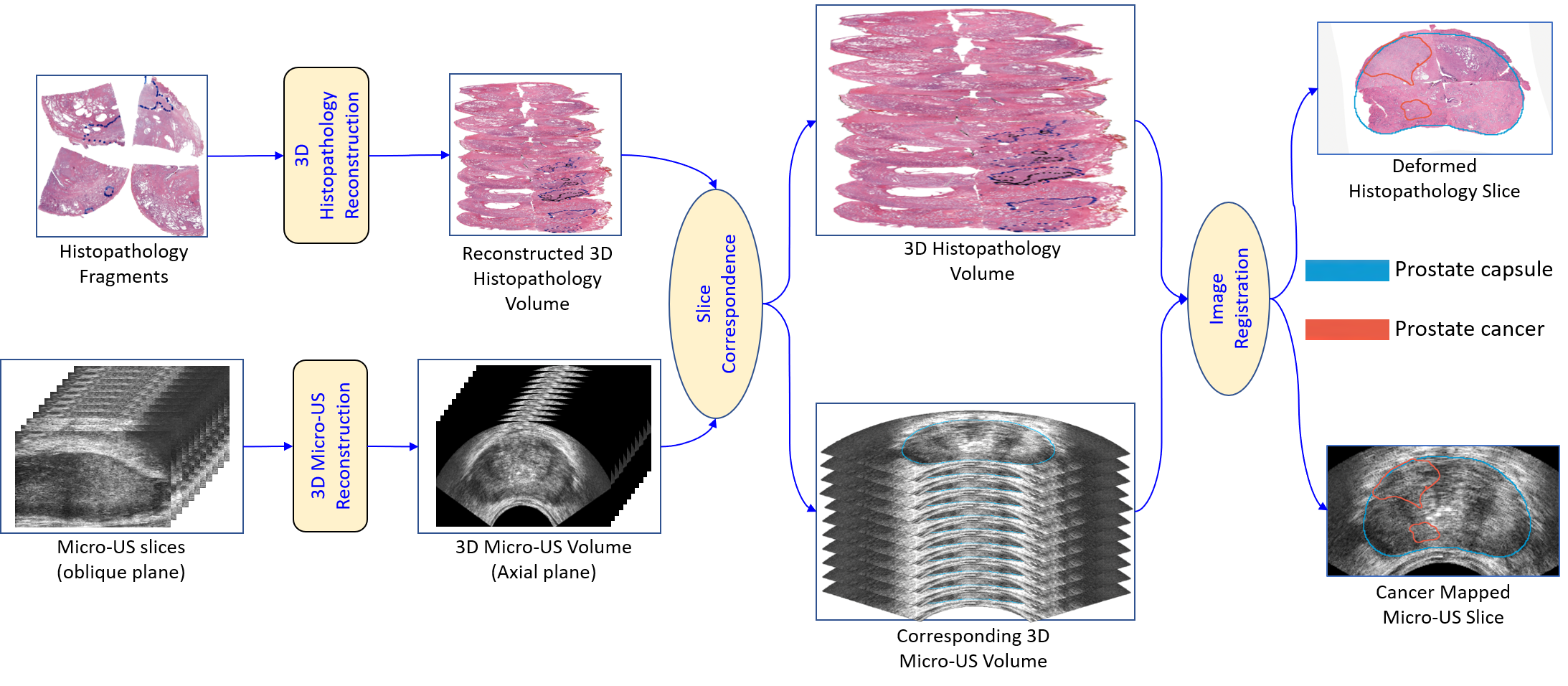}
\caption{An overview of our method for registering histopathology and micro-US images. 
Our first step involves stitching histology fragments to create pseudo-whole mount images, followed by constructing a 3D micro-US volume using micro-US images acquired in oblique planes.
Next, for each pseudo-whole mount histopathology image, we estimate the corresponding axial micro-US slice. 
Histopathology and micro-US image pairs are registered with affine and deformable registrations. 
The resulting transformation are used to map ground truth cancer labeling from histopathology onto micro-US images.
}
\label{Fig:techniqueoverview}
\end{figure*}

Our image pipeline involves several steps (see Figure \ref{Fig:techniqueoverview}), including reconstruction of pseudo-whole mount histopathology images and 3D micro-US volume, estimation of slice correspondences between histopathology images and micro-US images in the axial plane, and image registration of histopathology-micro-US image pairs.
In this section, we describe each step in detail.

\subsection{Dataset}
\label{sect:dataset}
 This study was approved by the University of Florida institutional review board.
With their informed consent, we included 18 patients who underwent radical prostatectomy at the University of Florida.
\subsubsection{Micro-ultrasound image acquisition}
\begin{figure}[!hbt]
\centering
\includegraphics[width=0.48\textwidth]{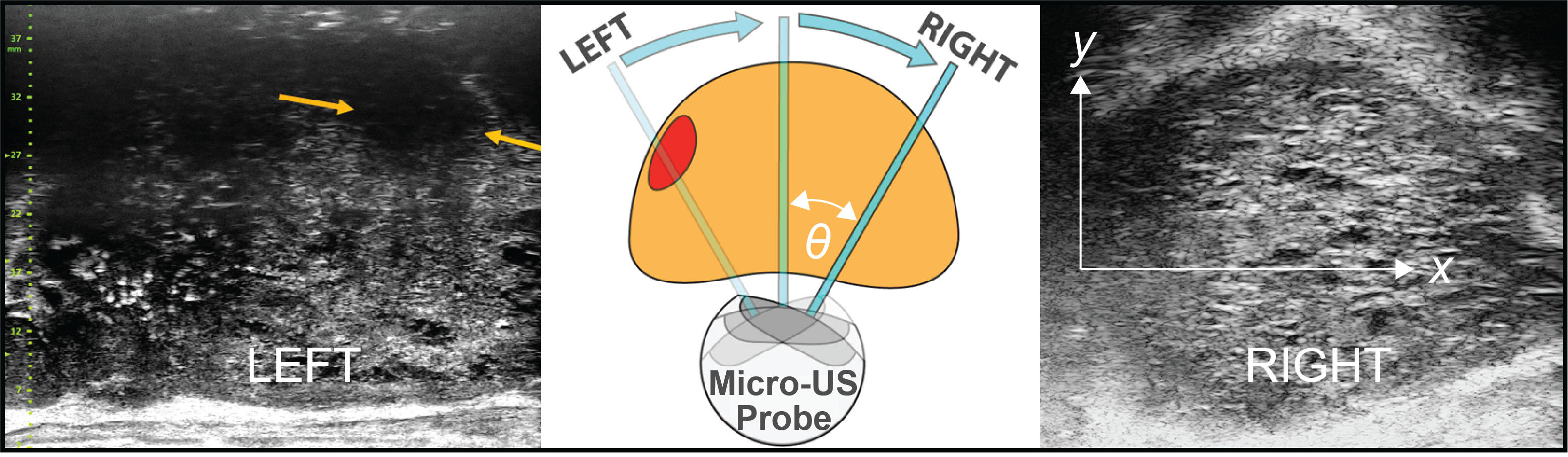}
\caption{The high-frequency micro-ultrasound is capable of imaging prostate cancer.  This prostate has a visible prostate cancer, indicated by the red region and yellow arrows.
A transrectal probe is inserted into the rectum and rotated to obtain micro-US images of the entire prostate at different angles. 
}
\label{Fig:Micro-USScanning}
\end{figure}

Each patient had a presurgical micro-US scan utilizing the ExactVu™ Micro-Ultrasound System (Exact Imaging, Markham, ON, Canada) \citep{ashouri2023micro}. Images were acquired in the oblique orientations by rotating the micro-US probe from the left to the right of the prostate (see Figure \ref{Fig:Micro-USScanning}).
Around 200 to 300 serial ultrasound images were recorded with each prostate sweep. The micro-US images have a fixed image size of $1372\times962$, with an isotropic in-plane resolution of 0.039 mm. The rotation angle associated with each image was stored by an accelerometer within the probe handle.

\subsubsection{Histopathology image acquisition}
The prostate that was surgically removed underwent formalin fixation, paraffin embedding, and sequential sectioning from the apex to the base in 3-mm slices perpendicular to the urethral axis. Since the size of the gross prostate sections exceeds the dimensions of our glass slides, we cut them into smaller tissue fragments and mount them into multiple slides (see Figure \ref{fig:fourquadrantsofhistoimage}(a)). The fragmented slices were stained using hematoxylin and eosin (H\&E) and digitized at 20x magnification using an Aperio Slide Scanner (Leica Biosystems, Buffalo Grove, IL), generating multiple histopathology fragments for each large section. The in-plane resolution of histopathology images is 0.0081 mm.
\subsubsection{Data annotation}
Prostate cancer was annotated on the histopathology images by an expert pathologist (SMF) with more than 10 years of experience.
Two imaging scientists (MI and WS) outlined the prostate capsule on reconstructed pseudo-whole mount histopathology images (see Section \ref{subsect:wholeslideimagereconstruction}) and reconstructed the axial micro-US images (see Section \ref{sec:reconstruction}).
Following the identification of slice correspondences between pseudo-whole mount histopathology and axial micro-US images (see Section \ref{sect:slicecorrespondence}), our imaging scientist (WS)  outlined the urethra and one anatomical landmark that was visible on both images, such as benign prostate hyperplasia nodules.
\subsection{Micro-ultrasound and histopathology image registration pipeline}
\label{sect:registrationoverview}
An overview of our registration approach is provided in Figure \ref{Fig:techniqueoverview}. The process begins with the reconstruction of pseudo-whole mount histopathology images and axial micro-US images. Subsequently, we manually establish correspondences between the histopathology and micro-US images. Using a conventional affine and deformable registration method, each pseudo-whole mount histopathology image is registered with its corresponding axial micro-US image. The resulting transformation is then utilized to map prostate cancer outlines from the histopathology images onto the micro-US images.
\subsubsection{Reconstruction of pseudo-whole mount histopathology images}
\label{subsect:wholeslideimagereconstruction}
\begin{figure}[!h]%
\centering
\includegraphics[width=0.5\textwidth]{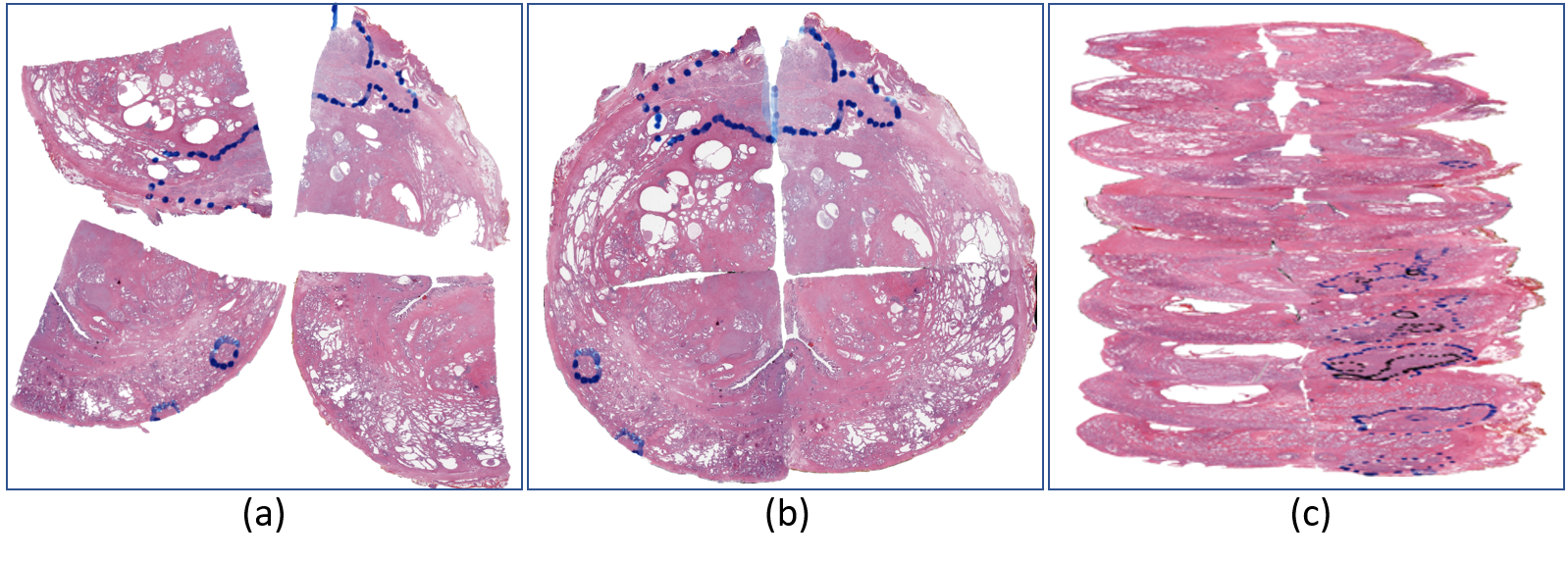}
    \caption{Initial Quadrant Images: The WSI is divided into four quadrants, each containing upper left, upper right, lower left, and lower right sections. These sections are not suitable for stitching in their initial orientation and placement. Therefore, geometric transformations are applied to properly orient them before stitching.}
    \label{fig:fourquadrantsofhistoimage}
\end{figure}

Obtaining whole-mount histological sections of large prostate glands while maintaining tissue integrity requires reconstruction. \citep{TothRobert2014HAis}. The size of these specimens often exceeds the capacity of a single glass slide, necessitating the division of the section into smaller fragments that can be distributed across multiple slides.  This approach is visually depicted in Figure \ref{fig:fourquadrantsofhistoimage}(a), where four histology fragments were obtained from a prostate section.

Before registration with micro-US images, it is necessary to stitch the histology fragments into pseudo-whole mount histopathology images.
In this study, we utilized the AutoSticher tool proposed by \citep{Ref25} due to its superior computational accuracy, time efficiency, and automatic procedure. 
During the histological preparation stage, the left side of the prostate was marked with black ink, while the right side was marked with blue ink. 
Using this information, we manually estimated the horizontal flipping and gross rotation angle for each histology fragment.
We then used AutoSticher to stitch histology quadrants into a pseudo-whole mount image that preserves tissue integrity (see Figure \ref{fig:fourquadrantsofhistoimage} (b)).
Since AutoSticher only accepts four fragments as input, for sections with only two fragments, we manually selected landmarks between the two fragments using the 3D Slicer tool \citep{fedorov20123d} to merge them into a pseudo-whole mount image. 
The stitching process was performed for each prostate tissue section  
to generate a 3D histology volume for each patient, as shown in Figure \ref{fig:fourquadrantsofhistoimage} (c). 

\subsubsection{Reconstruction of axial micro-ultrasound images}
\label{sec:reconstruction}
\begin{figure*}[!h]
\centering
\includegraphics[width=0.9\textwidth]{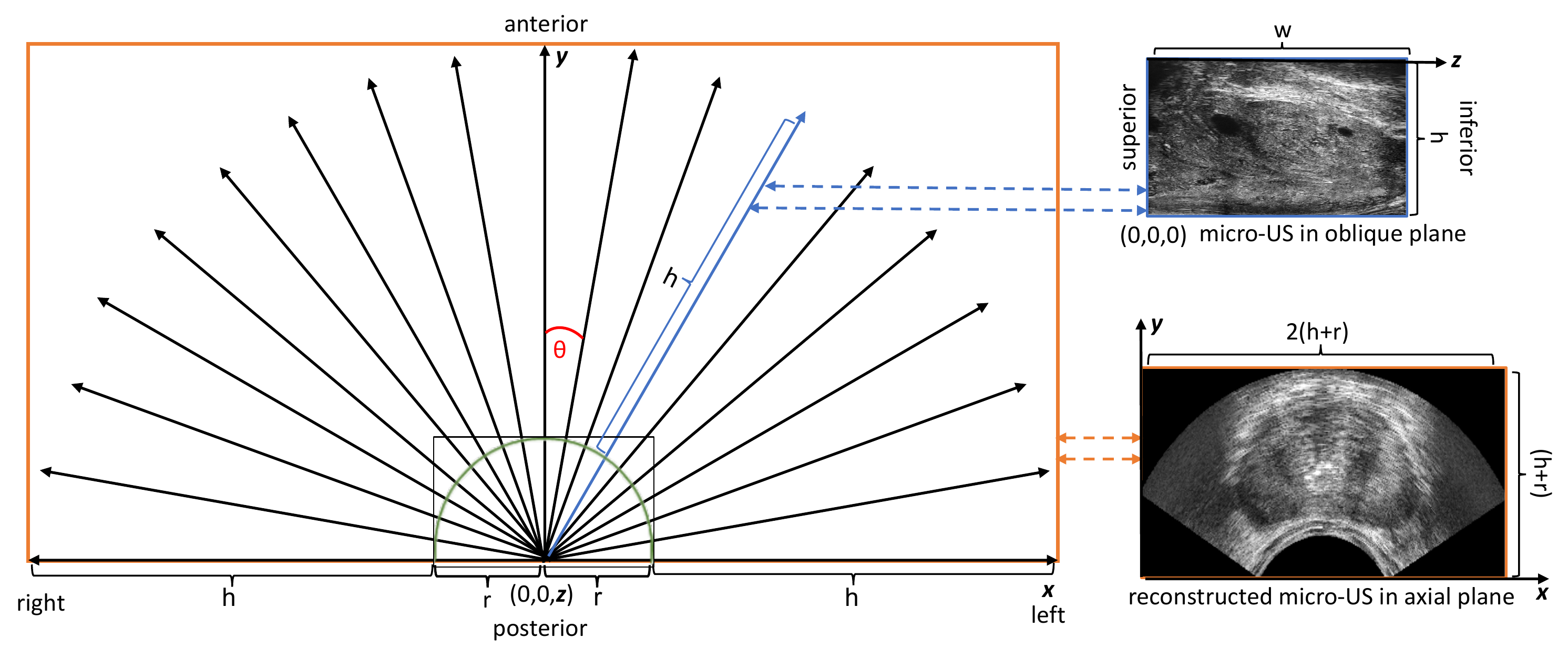}
 \caption{Acquisition of micro-US images in the oblique orientations and reconstruction of micro-US volume in the axial orientation. The radial arrows represent the acquired micro-US images at different rotation angles. We have defined the $x$, $y$, and $z$ directions for the reconstructed micro-US volume $V$. $h$ and $w$ are input micro-US image height and weight. $r$ is the radius of the probe. }
    \label{fig:microusconstruction}
\end{figure*}
To register with pseudo-whole mount histopathology images acquired in the axial orientation, it is necessary to reconstruct axial micro-US images from micro-US images acquired in the oblique planes.
The procedure of obtaining a micro-US scan of the prostate is similar to a standard TRUS procedure. 
The urologist uniformly rotates the probe from the left $(-90^{\circ})$ to the right $(90^{\circ})$ of the prostate within 20 to 30 seconds, with 10 frames recorded per second, producing an angular resolution of approximately $0.6-0.9^{\circ}$.

We have developed an algorithm for the reconstruction of a 3D micro-US volume $V$ from a set of $N$ micro-US images acquired at different angles.
We denote the rotation angle of the $i$-th micro-US image as $\theta_i$, and we refer to this image as $I_{\theta_i}$.
To construct a 3D micro-US volume, we first compute the physical size of the volume in each direction.
Let $r$ denote the probe's radius, and $h$ and $w$ denote the height and width of acquired micro-US images, respectively.
As shown in Figure \ref{fig:microusconstruction}, the physical size of the 3D micro-US volume from the left to the right ($x$-axis) is $S_{LR} = 2(h+r)$, from the posterior to the anterior ($y$-axis) it is $S_{PA} = h + r $, and from the superior to the inferior ($z$-axis) it is $S_{SI} = w$.
We define the origin (i.e., (0,0,0)) of the physical coordinate system of $V$ as the center of the probe positioned towards the base of the prostate. Figure \ref{fig:microusconstruction} shows that the coordinate of the lower left pixel in each input micro-US image is (0,0,0).
To determine the intensity value of the micro-US volume $V$ at the physical location $(x,y,z)$, we first estimate the rotation angle $\theta$ of the micro-US image that contains this voxel. 
The $x$ and $y$ coordinates determine the rotation angle, while the $z$ coordinate corresponds to the distance from the base of the prostate.
As shown in Figure \ref{fig:microusconstruction}, the tangent of the angle for $(x,y,z)$ is given by $tan\theta = \frac{x}{y}$.
Therefore, the micro-US image that contains the voxel $(x,y,z)$ has a rotation angle of $arctan(\frac{x}{y})$.
Next, we determine the 2D coordinates of the pixel in this micro-US image that corresponds to  $(x,y,z)$.
This pixel has a distance of $z$ from the left side of the 2D micro-US image and a distance of $\sqrt{x^2 + y^2}$ from the bottom of the 2D slice.
Hence, the intensity value of the pixel (x,y,z) is given by $I_{arctan(\frac{x}{y})}(z, h - \sqrt{x^2 + y^2})$. 

The users have the flexibility to specify the in-plane resolution $\sigma_i$ and the through-plane $\sigma_t$ resolution of the reconstructed volume.
The number of voxels in the left-right direction is $N_{LR} = \frac{S_{LR}}{\sigma_i}$, the posterior-anterior direction is $N_{PA} = \frac{S_{PA}}{\sigma_i}$, and the superior-inferior direction is $N_{SI} = \frac{S_{SI}}{\sigma_t}$.
In this paper, we chose $\sigma_i = 0.4 $ mm, and $\sigma_t = 1 $ mm.
We summarize the aforementioned approach in Algorithm \ref{Algo1}. 
\begin{algorithm}[th]
	\SetAlgoLined
	\KwIn{\\
 \Indp \Indp
 Micro-US images: $I_{\theta_1}, \cdots, I_{\theta_N}$\\
 Height and width of 2D micro-US : $h$, $w$
 Radius of the probe: $r$ \\
 In-plane resolution of 3D volume: $\sigma_i$ \\
 Through-plane resolution of 3D volume: $\sigma_t$ \\
 }
\KwOut{\\
 \Indp \Indp
 3D micro-ultrasound volume: $V$}

$\Theta = (\theta_1,\cdots, \theta_N)$
\Comment{angles of micro-US images}
 $N_{LR} = \frac{2(h+r)}{\sigma_i}$
 \Comment{$\#$  voxels in from left to right }
 $N_{PA} = \frac{h+r}{\sigma_i}$
 \Comment{$\#$  voxels from left to right}
  $N_{SI} = \frac{w}{\sigma_t}$
\Comment{$\#$ voxels from superior to inferior}

\Comment{$i,j,k$ are integers to represent voxel indices}

 \For{$i\gets1$ \KwTo $N_{LR}$  }{
 \For{$j\gets1$ \KwTo $N_{PA}$  }{
 \For{$k\gets1$ \KwTo $N_{SI}$  }{
\Comment{$x,y,z$ are the physical coordinates}
$x = \sigma_i \times i - (h+r)$ \\
$y = \sigma_i \times j $ \\
$z = \sigma_t \times k$ \\

 $\theta \gets arctan(\frac{x}{y})$ 
 \Comment{angle of this voxel}
$idx = \arg \min |\Theta - \theta|$ \\
\Comment{find index of micro-US image which has a rotation angle closest to $theta$}
$V[i,j,k] = I_{\theta{idx}}(z, h-\sqrt{(x^2 + y^2})$
\Comment{here we use $[]$ to denote image indices, and $()$ to denote physical coordinates.}
}}}
 \caption{3D micro-US reconstruction algorithm.}
 \label{Algo1}
\end{algorithm}

\subsubsection{Estimation of slice correspondences between histopathology and micro-ultrasound images}
\label{sect:slicecorrespondence}
\begin{figure}[!h]%
\centering
\includegraphics[width=0.5\textwidth]{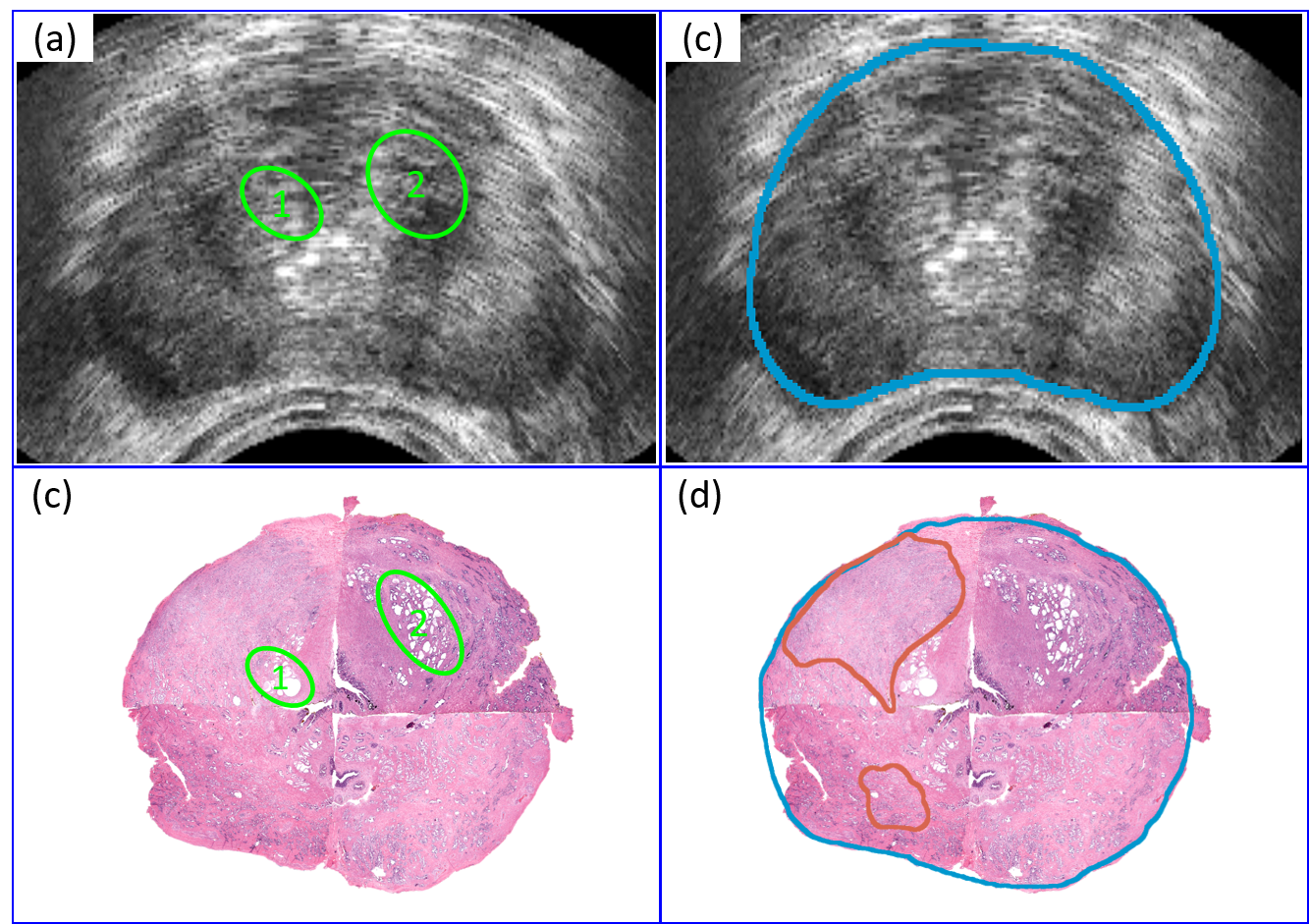}
    \caption{An example of paired micro-US (a) and histopathology images (c). To establish correspondence between (a) and (c), we relied on Region 1 and Region 2 as the prominent prostate features. Additionally, we manually segmented the prostate capsule on the micro-US image (b), and both the prostate capsule and prostate cancer on the corresponding histopathology image (d).}
    \label{fig:slice_correspondence}
\end{figure}

The next step is to establish slice correspondences between the reconstructed pseudo-whole mount histopathology images (captured every 3 mm) and the axial micro-US images (captured every 1 mm).
Since the distances between adjacent histopathology images and adjacent axial micro-US images are fixed, it is sufficient to find a single pair of histopathology and micro-US images that correspond to each other. To achieve this, we employ visual image features to identify the histopathology-micro-US image pair with the highest resemblance. Figure \ref{fig:slice_correspondence} displays the histopathology-micro-US image pair with the features with the strongest resemblance for a representative patient. In this case, the two areas outlined by yellow circles were utilized as the predominant features to determine the final estimation.


After establishing slice correspondences, we proceeded to manually outline the prostate capsule on both the histopathology images and their corresponding axial micro-US images. This prostate segmentation serves a dual purpose: facilitating the registration process and enabling evaluation of our registration approach. Additionally, we delineated the extent of prostate cancer on the histopathology images by referencing the cancer outlines provided by the pathologist on paper.

\subsubsection{ Two-stage image registration method}
\label{sect:segmentation}

The final step in our registration pipeline is to perform pairwise registrations of histopathology-micro-US image pairs. 
To accomplish this, we implemented a two-stage image registration approach consisting of an affine registration followed by a deformable registration.
During the registration process, a multi-resolution pyramid technique was employed, which comprised three layers with shrinking factors of 4, 8, and 2, and smoothing sigma values of 4, 2, and 1 pixel, respectively.
For the affine registration, we utilized the prostate masks as a guiding factor for optimization, employing the sum of square differences as the loss function.
For the deformable registration, we used free-form deformations, masked prostate images as the input, and Mattes mutual information as the loss function.
To optimize the registration, we employed a gradient descent optimizer with a learning rate of 0.2 and conducted 12 iterations per resolution layer for both the affine and deformable registrations.
All registrations were implemented in Python using the Simple ITK library \citep{lowekamp2013design}. 

\subsection{Evaluation metrics}
\label{sect:evaluationmetrics}
To measure the performance of our proposed methodology, we use four metrics: Dice coefficient, Hausdorff distance, urethra deviation, and landmark error. 
\subsubsection{Dice coefficient}
\label{sect:dicecoefficient}
We use the 2D Dice coefficient to evaluate the relative overlap between prostate segmentation on the fixed micro-US image and prostate segmentation on the deformed histopathology image. 
 The mean Dice coefficient for each patient is computed as follows:
\begin{equation}
Dice(I,J) = \frac{1}{K} \sum_{n=1}^{K}\left(2\times \frac{\left|I_n \bigcap J_n\right|}{|I_n| + |J_n|}\right),
\end{equation}
where $K$ is the number of pseudo-whole mount histopathology images, $I_n$ and $J_n$ are binary prostate segmentations for the $n$-th micro-US and registered histopathology images, and $\left| \cdot \right|$ represents the cardinality.
The Dice coefficient ranges from 0 (no overlap) to 1 (perfect overlap).

\subsubsection{Hausdorff distance}
\label{sect:hausdorffdistance}
We use Hausdorff distance \citep{HausdorffDistance} to measure the distance between prostate boundaries outlined on the micro-US and aligned histopathology images. 
Equation (\ref{eq:hausdorffdistance}) gives the mean Hausdorff distance (HD) for each patient:
\begin{equation}
\label{eq:hausdorffdistance}
\begin{aligned}
&HD(I,J) = \\ 
&\frac{1}{K}\sum_{n=1}^{K}\left(\!\!\max \!\!\left\{ \sup_{i\in I_n} \inf_{b \in J_n} d(i, b), \sup_{i\in I_n} \inf_{b \in J_n} d(i, b) \!\!\right\}\!\!\right)
\end{aligned}
\end{equation}
where $\sup$ and $\inf$ denote the supremum (least upper bound) and infimum (greatest lower bound) operators, respectively; $K$ represents the number of 2D histopathology-micro-US image pairs; $I_n$ and $J_n$ are binary prostate segmentations.

\subsubsection{Urethra deviation}
\label{sect:urethradeviation}
One metric used to assess the alignment of interior prostate features is urethra deviation, which measures the Euclidean distance between the centers of mass of the urethra on the micro-US and deformed histopathology images. 
We define urethra deviation (UD) as follows:
\begin{equation}
\label{eq:urethradeviation}
UD\left(I, J\right) = \frac{1}{K}\sum_{n = 1}^{K}\left(\sqrt{(x_n^m - x_n^h)^2+(y_n^m - y_n^h)^2}\right),
\end{equation}
where $(x_n^m, y_n^m)$ and $(x_n^h, y_n^h)$ are the coordinates of the center of mass of the urethra in $n$-th micro-US image and the corresponding deformed histopathology image.

\subsubsection{Landmark error}
\label{sect:landmarkerror}
In addition to assessing the urethra deviation, we use landmark error to estimate the alignment of anatomical prostate features. 
For each pair of micro-US and histopathology images, we select one anatomical landmark (e.g., benign prostate hyperplasia) in both images. The landmark error (LE) is defined as follows:
\begin{equation}
\label{eq:landmarkerror}
LE\left(I, J\right) = \frac{1}{K}\sum_{n = 1}^{K}\left(\sqrt{(x_n^m - x_n^h)^2+(y_n^m - y_n^h)^2}\right),
\end{equation}
where $K$ is the number of 2D; $(x_n^m, y_n^m)$  and $(x_n^h, y_n^h)$ are the coordinates of the landmark centers in the $n$-th micro-US image and deformed histopathology image.

\section{Results}
\label{sec:results}

\subsection{Qualitative results}
\begin{figure}[!hbt]
\centering
\includegraphics[width=0.48\textwidth]{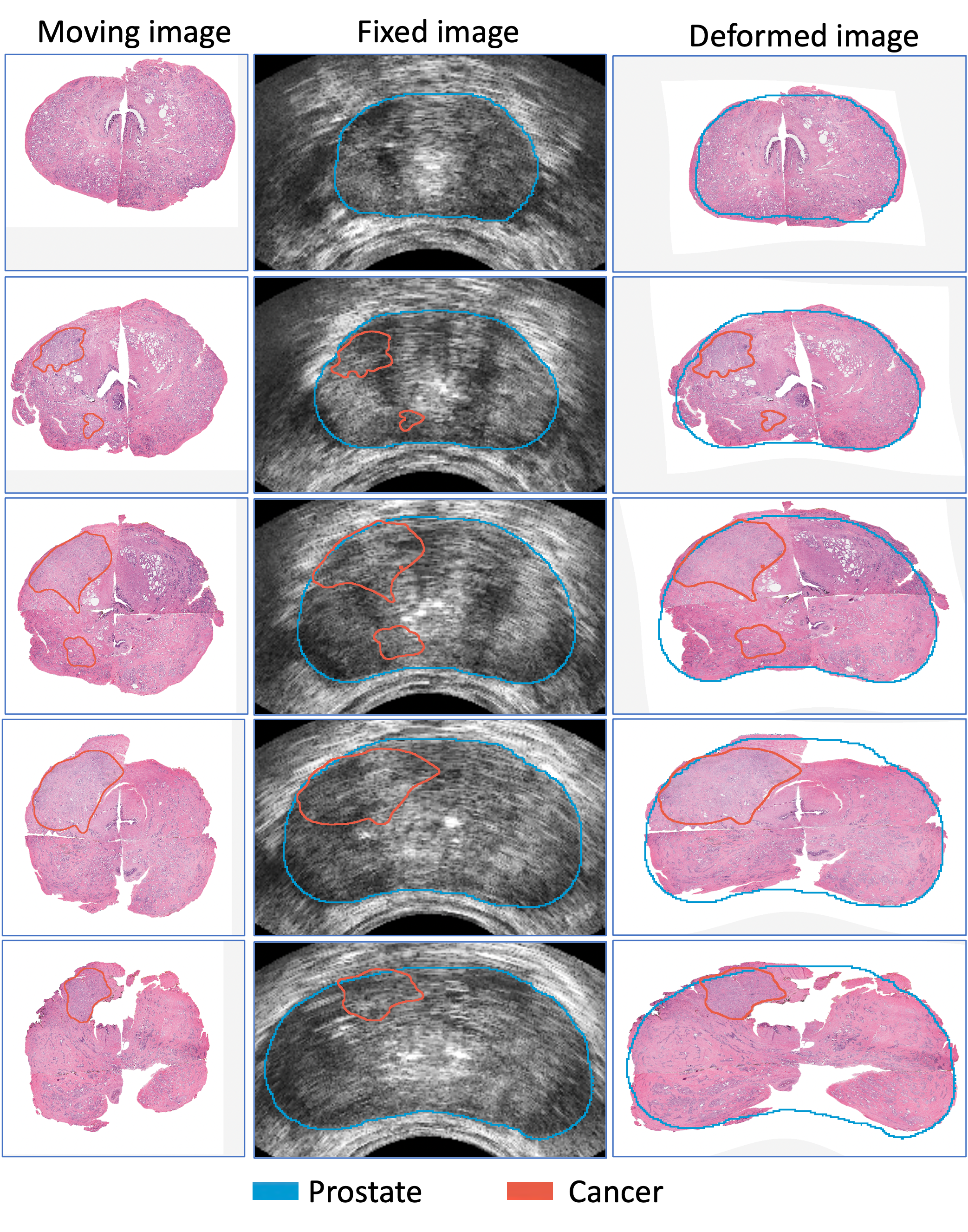}
\caption{ Qualitative registration results of a representative patient from the apex to the base of the prostate. 
First row: moving histopathology images with pathologist-annotated prostate cancer outlines.
Second row: fixed micro-US images with manual prostate segmentation as well as cancer outlines mapped from histopathology images.
Third row: registered (deformed) histopathology images overlaid with prostate segmentation on the fixed images. 
}
\label{Fig:QualitativeResults}
\end{figure}

We successfully applied our registration pipeline to register the pseudo-whole mount histopathology and axial micro-US images of 18 patients who underwent radical prostatectomy. In our approach, we utilized the histopathology images as the moving images and the corresponding axial micro-US images as the fixed images.
Figure \ref{Fig:QualitativeResults} showcases the registration results for a representative patient, spanning from the apex to the base of the prostate. In the first row of Figure \ref{Fig:QualitativeResults}, we have delineated the extent of prostate cancer on the moving histopathology images. 
For enhanced visualization, we have also outlined the prostate capsule on the fixed images, as demonstrated in the second row of Figure \ref{Fig:QualitativeResults}.
The third row  of Figure \ref{Fig:QualitativeResults} presents the deformed histopathology images, highlighting the effectiveness of our registration pipeline in two key aspects.
First, the prostate segmentation on the micro-US images (indicated by blue outlines) aligns well with the prostate boundaries of the deformed histopathology images.
Second, important anatomical features such as the urethra and cancer are also well-aligned. 
This alignment of the prostate and anatomical landmarks serves as validation for the accuracy and reliability of our registration methodology.
The precise registration of radiology-pathology images enables us to accurately map the ground truth cancer outlines from histopathology onto micro-US images, as displayed in the second row of Figure \ref{Fig:QualitativeResults}.

\subsection{Quantitative results}
We conducted a quantitative evaluation to assess the alignment of prostate boundaries using the Dice coefficient and Hausdorff distance, and to evaluate the alignment of anatomical features using the urethra deviation and landmark error.
Table \ref{tab:quantativeresults} presents the results of these metrics for our dataset consisting of 18 cases.
The mean Dice coefficient of 0.965 and the mean Hausdorff distance of 1.77 mm illustrate the high accuracy achieved in image registration near the prostate boundaries.
Furthermore, the mean urethra deviation of 2.20 mm and the mean landmark error of 2.71 mm are comparable to the results obtained by a state-of-the-art MRI-histopathology registration approach \citep{shao2021prosregnet}, which achieved a urethra deviation of 2.37 mm and a landmark error of 2.68 mm.
These findings demonstrate the feasibility and accuracy of our micro-US and histopathology image registration methodology.

\begin{table}[!h]                       
	\centering
	\renewcommand{\arraystretch}{1.3} 
	\begin{tabular}{c|c |c|c|c}
		\toprule[1.5pt]
  \multirow{3}{*}{Case No.} & Dice & Hausdorff & Urethra & Landmark \\
  & Coeff& Distance & Deviation & Error\\
  &  & (mm) & (mm) & (mm) \\
		\midrule[1.5pt]
		1   &0.96	 &  1.538	&   4.057	&   4.711 \\
        2   &0.961   &	1.479   &	1.758   &	4.141\\
        3   &0.972   &	1.324   &	2.296   &	4.542\\
        4   &0.973   &	2.386   &	1.508   &	2.211\\
        5   &0.964   &	1.105   &	2.129   &	2.264\\
        6   &0.968   &	1.755   &	1.544   &	2.389\\
        7   &0.968   &	1.519   &	1.609   &	2.85\\
        8   &0.967   &	1.462   &	1.431   &	0.984\\
        9   &0.977   &	0.959   &	2.438   &	3.038\\
        10   &0.952   &	2.557   &	1.552   &	1.828\\
        11   &0.945   &	2.541   &	1.255   &	2.737\\
        12   &0.953   &	3.267   &	2.109   &	3.442\\
        13   &0.973   &	2.003   &	2.609   &	2.261\\
        14   &0.956   &	2.222   &	3.749   &	2.256\\
        15   &0.965   &	1.599   &	2.017   &	1.728\\
        16   &0.973   &	1.232   &	2.183   &	2.856\\
        17   &0.978   &	1.296   &	2.515   &	2.361\\
        18   &0.962   &	1.549   &	2.83   &	2.091\\
        \hline
        Average   &\textbf{0.965}   &	\textbf{1.766}   &	\textbf{2.2}   &	\textbf{2.705}\\
		\bottomrule[1.5pt]
	\end{tabular}
	\caption{Quantitative evaluation of the registration results of 18 patients.}
	\label{tab:quantativeresults}                           	
\end{table}

\section{Discussion}
\label{sect:discussion}
\subsection{Clinical implications}
Micro-US is a promising diagnostic modality with numerous advantages over MRI, including low cost, real-time imaging, and superior sensitivity. However, it remains a relatively new technology with the U.S. Food and Drug Administration approving it in 2016; thus, many clinicians lack experience with interpreting micro-US images.  The PRI-MUS score
 (prostate risk identification using micro-US) was developed to aid identification of micro-US features most indicative of cancer (Ghai et al., 2016).  However, PRI-MUS was developed using only 100 biopsy patients.  Because there are large areas of unknown tissue in biopsy patients, PRI-MUS has no quantitative features.  Future iterations of the PRI-MUS score will benefit from registered surgical pathology as ground truth. 
 
\begin{figure}[!h]%
\centering
\includegraphics[width=0.48\textwidth]{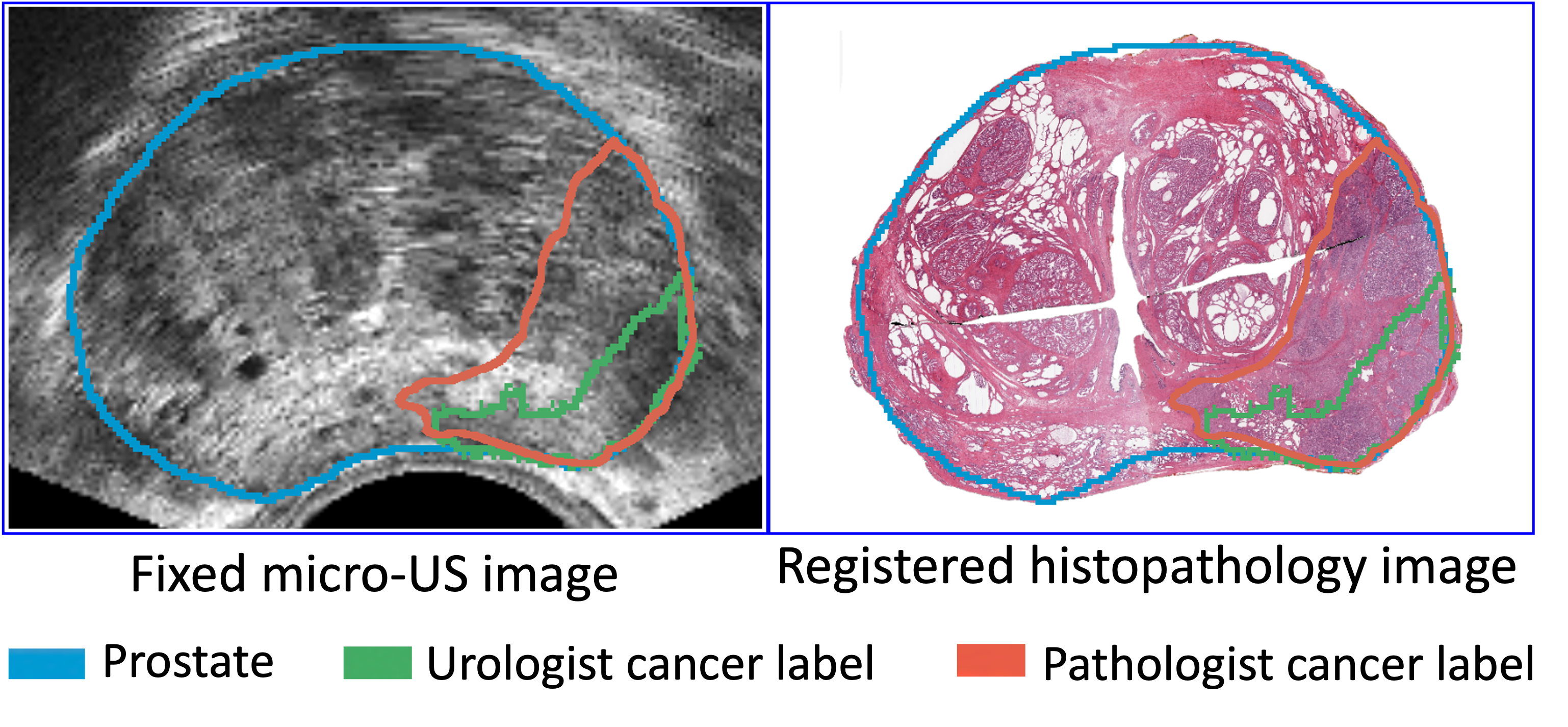}
    \caption{ 
    Urologists tend to under-segment the extent of prostate cancer on micro-US images.
    Here we show an example of registered micro-US and histopathology images. 
   Blue outline: manual prostate segmentation on micro-US.
   Green outline: urologist cancer outline on micro-US; the urologist (WB) referred to pathology results from biopsy cores during the annotation.
   Orange outline: pathologist cancer outline deformed by image registration.
   }
\label{fig:urologistpathologistcomparison}
\end{figure}

It is also worth noting that cancer outlines drawn by urologists in the biopsy cohort often underestimate the extent of prostate cancer when compared to the ground truth pathologist cancer outlines on whole-mount histopathology images (see Figure \ref{fig:urologistpathologistcomparison}). While tumor underestimation also limits MRI, the extent of underestimation is currently unidentified, thus limiting treatment to whole-gland therapy.  To overcome this limitation, we have implemented a radiology-pathology image registration pipeline that accurately maps the ground truth cancer outlines from histopathology onto micro-US images.

Finally, this pipeline facilitates the development of a large dataset of micro-US scans with ground truth cancer outlines. Such a dataset can be utilized to train both urologists and machine learning models to identify cancer lesions on micro-US images, thereby improving the accuracy of cancer detection in micro-US imaging.

\subsection{Limitations of our study}
This study has a few limitations that should be noted. 
First, as a proof-of-concept study, our image registration pipeline was validated using a relatively small dataset comprising 18 patients who had a radical prostatectomy. Additionally, our pipeline requires several manual steps, such as pseudo-whole mount histopathology image reconstruction, prostate segmentation on micro-US and histopathology images, and estimation of micro-US-histopathology slice correspondences.
In our future work, we will expand our dataset by collecting data from a larger cohort of patients. 
Furthermore, we will explore the development of deep learning techniques to streamline the registration pipeline by automating the manual steps. 
These advancements will allow for a more efficient and robust image registration process.

\section{Conclusion}
\label{sect:conclusion}
In this paper, we have introduced a semi-automated pipeline that successfully registers micro-US and pseudo-whole mount histopathology images of the prostate. 
This study serves as a proof-of-concept, demonstrating the feasibility of accurately aligning micro-US and histopathology images acquired at different orientations.
Additionally, we have curated a unique dataset comprising micro-US and aligned histopathology images, along with ground truth cancer outlines. 
Through our work, we aim to contribute to the advancement and wider adoption of micro-US as a promising technology for enhancing prostate cancer diagnosis.
\printcredits

\section*{Conflict of interest statement}
The authors have no conflict of interest to disclose.

\section*{Declaration of generative AI in scientific writing}
During the preparation of this work, the authors used the ChatGPT-3.5 model in order to improve the readability and language of our paper. After using this tool/service, the authors reviewed and edited the content as needed and take full responsibility for the content of the publication.

\section*{Acknowledgments}
This work was supported by the Department of Medicine and the Intelligent Critical Care Center at the University of Florida College of Medicine.
We would like to express our gratitude to Jessica Kirwan for editing the language of this paper.

\bibliographystyle{apalike}


\bibliography{biblioentropy} 

\begin{thebibliography}{}

\bibitem[Ashouri et~al., 2023]{ashouri2023micro}
Ashouri, R., Nguyen, B., Archer, J., Crispen, P., O’Malley, P., Su, L.-M.,
  Grajo, J., Falzarano, S.~M., Acar, Y., Lizdas, D., et~al. (2023).
\newblock Micro-ultrasound guided transperineal prostate biopsy: A clinic-based
  procedure.
\newblock {\em JoVE (Journal of Visualized Experiments)}, (192):e64772.

\bibitem[Avolio et~al., 2021]{Ref10}
Avolio, P.~P., Lughezzani, G., Paciotti, M., Maffei, D., Uleri, A., Frego, N.,
  Hurle, R., Lazzeri, M., Saita, A., Guazzoni, G., Casale, P., and Buffi, N.~M.
  (2021).
\newblock The use of 29 mhz transrectal micro-ultrasound to stratify the
  prostate cancer risk in patients with pi-rads iii lesions at multiparametric
  mri: A single institutional analysis.
\newblock {\em Urologic Oncology: Seminars and Original Investigations},
  39(12):832.e1--832.e7.

\bibitem[Chappelow et~al., 2011]{Ref17}
Chappelow, J., Bloch, B.~N., Rofsky, N., Genega, E., Lenkinski, R., DeWolf, W.,
  and Madabhushi, A. (2011).
\newblock Elastic registration of multimodal prostate mri and histology via
  multiattribute combined mutual information.
\newblock {\em Medical Physics}, 38(4):2005--2018.

\bibitem[Dias et~al., 2022]{Ref6}
Dias, A.~B., O’Brien, C., Correas, J.-M., and Ghai, S. (2022).
\newblock Multiparametric ultrasound and micro-ultrasound in prostate cancer: a
  comprehensive review.
\newblock {\em The British Journal of Radiology}, 95(1131):20210633.

\bibitem[Fedorov et~al., 2012]{fedorov20123d}
Fedorov, A., Beichel, R., Kalpathy-Cramer, J., Finet, J., Fillion-Robin, J.-C.,
  Pujol, S., Bauer, C., Jennings, D., Fennessy, F., Sonka, M., et~al. (2012).
\newblock 3d slicer as an image computing platform for the quantitative imaging
  network.
\newblock {\em Magnetic resonance imaging}, 30(9):1323--1341.

\bibitem[Galv{\'a}n et~al., 2023]{Ref2}
Galv{\'a}n, G.~C., Das, S., Daniels, J.~P., Friedrich, N.~A., and Freedland,
  S.~J. (2023).
\newblock Working hard or hardly working? a brief commentary of latest research
  on exercise and prostate cancer.
\newblock {\em Prostate Cancer and Prostatic Diseases}, pages 1--2.

\bibitem[Klotz et~al., 2020]{Ref8}
Klotz, L., Lughezzani, G., Maffei, D., Sánchez, A., Pereira, J.~G., Staerman,
  F., Cash, H., Luger, F., Lopez, L., Sanchez-Salas, R., Abouassally, R.,
  Shore, N.~D., and Eure, G. (2020).
\newblock Comparison of micro-ultrasound and multiparametric magnetic resonance
  imaging for prostate cancer: A multicenter, prospective analysis.
\newblock {\em Canadian Urological Association Journal}, 15(1):E11--6.

\bibitem[Li et~al., 2020]{Ref1}
Li, J., Xu, C., Lee, H.~J., Ren, S., Zi, X., Zhang, Z., Wang, H., Yu, Y., Yang,
  C., Gao, X., et~al. (2020).
\newblock A genomic and epigenomic atlas of prostate cancer in asian
  populations.
\newblock {\em Nature}, 580(7801):93--99.

\bibitem[Loeb et~al., 2011]{loeb2011complications}
Loeb, S., Carter, H.~B., Berndt, S.~I., Ricker, W., and Schaeffer, E.~M.
  (2011).
\newblock Complications after prostate biopsy: data from seer-medicare.
\newblock {\em The Journal of urology}, 186(5):1830--1834.

\bibitem[Lowekamp et~al., 2013]{lowekamp2013design}
Lowekamp, B.~C., Chen, D.~T., Ib{\'a}{\~n}ez, L., and Blezek, D. (2013).
\newblock The design of simpleitk.
\newblock {\em Frontiers in neuroinformatics}, 7:45.

\bibitem[Lughezzani et~al., 2019]{Ref9}
Lughezzani, G., Saita, A., Lazzeri, M., Paciotti, M., Maffei, D., Lista, G.,
  Hurle, R., Buffi, N.~M., Guazzoni, G., and Casale, P. (2019).
\newblock Comparison of the diagnostic accuracy of micro-ultrasound and
  magnetic resonance imaging/ultrasound fusion targeted biopsies for the
  diagnosis of clinically significant prostate cancer.
\newblock {\em European Urology Oncology}, 2(3):329--332.

\bibitem[Moskalik et~al., 1997]{Ref21}
Moskalik, A., Carson, P., Rubin, J., Fowlkes, J., Wojno, K., and Bree, R.
  (1997).
\newblock 3d registration of ultrasound with histology in the prostate.
\newblock In {\em 1997 IEEE Ultrasonics Symposium Proceedings. An International
  Symposium (Cat. No.97CH36118)}, volume~2, pages 1397--1400 vol.2.

\bibitem[Penzias et~al., 2016]{Ref25}
Penzias, G., Janowczyk, A., Singanamalli, A., Rusu, M., Shih, N., Feldman, M.,
  Stricker, P.~D., Delprado, W., Tiwari, S., Böhm, M., Haynes, A.-M., Ponsky,
  L., Viswanath, S., and Madabhushi, A. (2016).
\newblock Autostitcher: An automated program for efficient and robust
  reconstruction of digitized whole histological sections from tissue
  fragments.
\newblock {\em Scientific Reports}, (29906).

\bibitem[Porter et~al., 2001]{Ref22}
Porter, B., Taylor, L., Baggs, R., di~Sant'Agnese, A., Nadasdy, G., Pasternack,
  D., Rubens, D., and Parker, K. (2001).
\newblock Histology and ultrasound fusion of excised prostate tissue using
  surface registration.
\newblock In {\em 2001 IEEE Ultrasonics Symposium. Proceedings. An
  International Symposium (Cat. No.01CH37263)}, volume~2, pages 1473--1476
  vol.2.

\bibitem[Schalk et~al., 2016]{Ref11}
Schalk, S.~G., Postema, A., Saidov, T.~A., Demi, L., Smeenge, M., {de la
  Rosette}, J.~J., Wijkstra, H., and Mischi, M. (2016).
\newblock 3d surface-based registration of ultrasound and histology in prostate
  cancer imaging.
\newblock {\em Computerized Medical Imaging and Graphics}, 47:29--39.

\bibitem[Schömig-Markiefka et~al., 2021]{Ref18}
Schömig-Markiefka, B., Pryalukhin, A., Hulla, W., Bychkov, A., Fukuoka, J.,
  Madabhushi, A., Achter, V., Nieroda, L., Büttner, R., Quaas, A., and
  Tolkach, Y. (2021).
\newblock Quality control stress test for deep learning-based diagnostic model
  in digital pathology.
\newblock {\em Modern Pathology}, 34(12):2098--2108.

\bibitem[Shao et~al., 2021]{shao2021prosregnet}
Shao, W., Banh, L., Kunder, C.~A., Fan, R.~E., Soerensen, S.~J., Wang, J.~B.,
  Teslovich, N.~C., Madhuripan, N., Jawahar, A., Ghanouni, P., et~al. (2021).
\newblock {ProsRegNet}: a deep learning framework for registration of {MRI} and
  histopathology images of the prostate.
\newblock {\em Medical image analysis}, 68:101919.

\bibitem[Siegel et~al., 2023]{Ref3}
Siegel, R.~L., Miller, K.~D., Wagle, N.~S., and Jemal, A. (2023).
\newblock Cancer statistics, 2023.
\newblock {\em CA: A Cancer Journal for Clinicians}, 73(1):17--48.

\bibitem[Sood et~al., 2021]{sood20213d}
Sood, R.~R., Shao, W., Kunder, C., Teslovich, N.~C., Wang, J.~B., Soerensen,
  S.~J., Madhuripan, N., Jawahar, A., Brooks, J.~D., Ghanouni, P., et~al.
  (2021).
\newblock {3D Registration of pre-surgical prostate MRI and histopathology
  images via super-resolution volume reconstruction}.
\newblock {\em Medical Image Analysis}, 69:101957.

\bibitem[Sountoulides et~al., 2021]{Ref7}
Sountoulides, P., Pyrgidis, N., Polyzos, S.~A., Mykoniatis, I., Asouhidou, E.,
  Papatsoris, A., Dellis, A., Anastasiadis, A., Lusuardi, L., and
  Hatzichristou, D. (2021).
\newblock Micro-ultrasound–guided vs multiparametric magnetic resonance
  imaging-targeted biopsy in the detection of prostate cancer: A systematic
  review and meta-analysis.
\newblock {\em Journal of Urology}, 205(5):1254--1262.

\bibitem[Toth et~al., 2014]{TothRobert2014HAis}
Toth, R., Shih, N., Tomaszewski, J., Feldman, M., Kutter, O., Yu, D., et~al.
  (2014).
\newblock Histostitcher™: An informatics software platform for reconstructing
  whole-mount prostate histology using the extensible imaging platform
  framework.
\newblock {\em Journal of Pathology Informatics}, 5(1):8--8.

\bibitem[Wang et~al., 2020]{HausdorffDistance}
Wang, Z., Wang, E., and Zhu, Y. (2020).
\newblock Image segmentation evaluation: a survey of methods.
\newblock {\em Artificial Intelligence Review}, 53:5637–5674.

\bibitem[Ward et~al., 2012]{Ref19}
Ward, A.~D., Crukley, C., McKenzie, C.~A., Montreuil, J., Gibson, E.,
  Romagnoli, C., Gomez, J.~A., Moussa, M., Chin, J., Bauman, G., and Fenster,
  A. (2012).
\newblock Prostate: Registration of digital histopathologic images to in vivo
  mr images acquired by using endorectal receive coil.
\newblock {\em Radiology}, 263(3):856--864.

\bibitem[Zamboglou et~al., 2021]{Ref20}
Zamboglou, C., Kramer, M., Kiefer, S., Bronsert, P., Ceci, L., Sigle, A.,
  Schultze-Seemann, W., Jilg, C.~A., Sprave, T., Fassbender, T.~F., Nicolay,
  N.~H., Ruf, J., Benndorf, M., Grosu, A.~L., and Spohn, S. K.~B. (2021).
\newblock The impact of the co-registration technique and analysis methodology
  in comparison studies between advanced imaging modalities and
  whole-mount-histology reference in primary prostate cancer.
\newblock {\em Scientific Reports}, 5839.

\end{thebibliography}

\end{document}